\documentclass{article}
\usepackage[final, nonatbib]{main}

\usepackage[utf8]{inputenc} 
\usepackage[T1]{fontenc}    
\usepackage{hyperref}       
\usepackage{url}            
\usepackage{booktabs}       
\usepackage{amsfonts}       
\usepackage{nicefrac}       
\usepackage{microtype}      
\usepackage{xcolor}         
\usepackage{makecell}
\usepackage[accsupp]{axessibility}  
\usepackage{bm}
\usepackage{amsmath}
\usepackage{float}
\usepackage{subfig}
\usepackage{array}
\usepackage{siunitx}
\usepackage{color}
\usepackage{colortbl}
\usepackage{graphicx}
\usepackage{enumitem}
\usepackage{tablefootnote}
\usepackage{bbding}

\def\modelname{\texttt{RegionSpot}}
\def\modelnamelite{\texttt{RegionSpot-\texttt{Lite}}}
\def\modelnamepro{\texttt{RegionSpot-\texttt{Pro}}}
\def\modelnameproplus{\texttt{RegionSpot-\texttt{Pro{\scriptsize ↑336}}}}

\definecolor{cyan}{cmyk}{.3,0,0,0}
\definecolor{hr}{gray}{0.5}
\definecolor{hg}{gray}{0.8}
\def\onedot{.}
\def\eg{{\em e.g}\onedot} 
\def\ie{{\em i.e}\onedot}

\def\vs{{\em vs}\onedot}

\definecolor{cyan}{cmyk}{.3,0,0,0}
\definecolor{hr}{gray}{0.5}

\newlength\savewidth\newcommand\shline{\noalign{\global\savewidth\arrayrulewidth
		\global\arrayrulewidth 1pt}\hline\noalign{\global\arrayrulewidth\savewidth}}

\newcolumntype{*}{>{\global\let\currentrowstyle\relax}}
\newcolumntype{^}{>{\currentrowstyle}}

\newcolumntype{x}[1]{>{\centering\arraybackslash}p{#1pt}}
\newcolumntype{y}[1]{>{\raggedright\arraybackslash}p{#1pt}}
\newcolumntype{z}[1]{>{\raggedleft\arraybackslash}p{#1pt}}

\title{Recognize Any Regions
}

\author{%
  Haosen Yang\textsuperscript{1}\thanks{This work was performed when Haosen Yang worked as an intern at ByteDance.}\footnotemark[1] \quad
  Chuofan Ma\textsuperscript{2} \quad
  Bin Wen\textsuperscript{3} \quad
  Yi Jiang\textsuperscript{3}\thanks{Corresponding authors}\footnotemark[2] \quad
  Zehuan Yuan\textsuperscript{3} \quad
  Xiatian Zhu\textsuperscript{1}\footnotemark[2] \\
  \\
  \textsuperscript{1}University of Surrey \quad
  \textsuperscript{2}The University of Hong Kong \quad
  \textsuperscript{3}ByteDance \\
  \\
}

\begin{document}

\maketitle

\begin{abstract}
\label{sec:abb}
Understanding the semantics of individual regions or patches of unconstrained images, such as open-world object detection, remains a critical yet challenging task in computer vision. Building on the success of powerful image-level vision-language (ViL) foundation models like CLIP, recent efforts have sought to harness their capabilities by either training a contrastive model from scratch with an extensive collection of region-label pairs or aligning the outputs of a detection model with image-level representations of region proposals. Despite notable progress, these approaches are plagued by computationally intensive training requirements, susceptibility to data noise, 
and deficiency in contextual information.
To address these limitations, we explore the synergistic potential of off-the-shelf foundation models, leveraging their respective strengths in localization and semantics. We introduce a novel, generic, and efficient architecture, named \textit{\modelname{}}, designed to integrate position-aware localization knowledge from a localization foundation model (e.g., SAM) with semantic information from a ViL model (e.g., CLIP). To fully exploit pretrained knowledge while minimizing training overhead, we keep both foundation models frozen, focusing optimization efforts solely on a lightweight attention-based knowledge integration module.
Extensive experiments in open-world object recognition show that our \textit{\modelname{}} achieves significant performance gain over prior alternatives, along with substantial computational savings (e.g., training our model with 3 million data in a single day using 8 V100 GPUs). 
{\modelname{} outperforms GLIP-L by 2.9 in mAP on LVIS val set,
with an even larger margin of 13.1 AP for more challenging and rare categories, and a 2.5 AP increase on ODinW.
Furthermore, it exceeds GroundingDINO-L by 11.0 AP for rare categories on the LVIS minival set. Code is available at: \href{https://github.com/Surrey-UPLab/Recognize-Any-Regions}{\texttt{https://github.com/Surrey-UPLab/Recognize-Any-Regions}}
}
\end{abstract}

\section{Introduction}
\label{sec:intro}
\begin{figure}[]
\centering
\vspace{-5pt}
\includegraphics[width=0.9\linewidth]{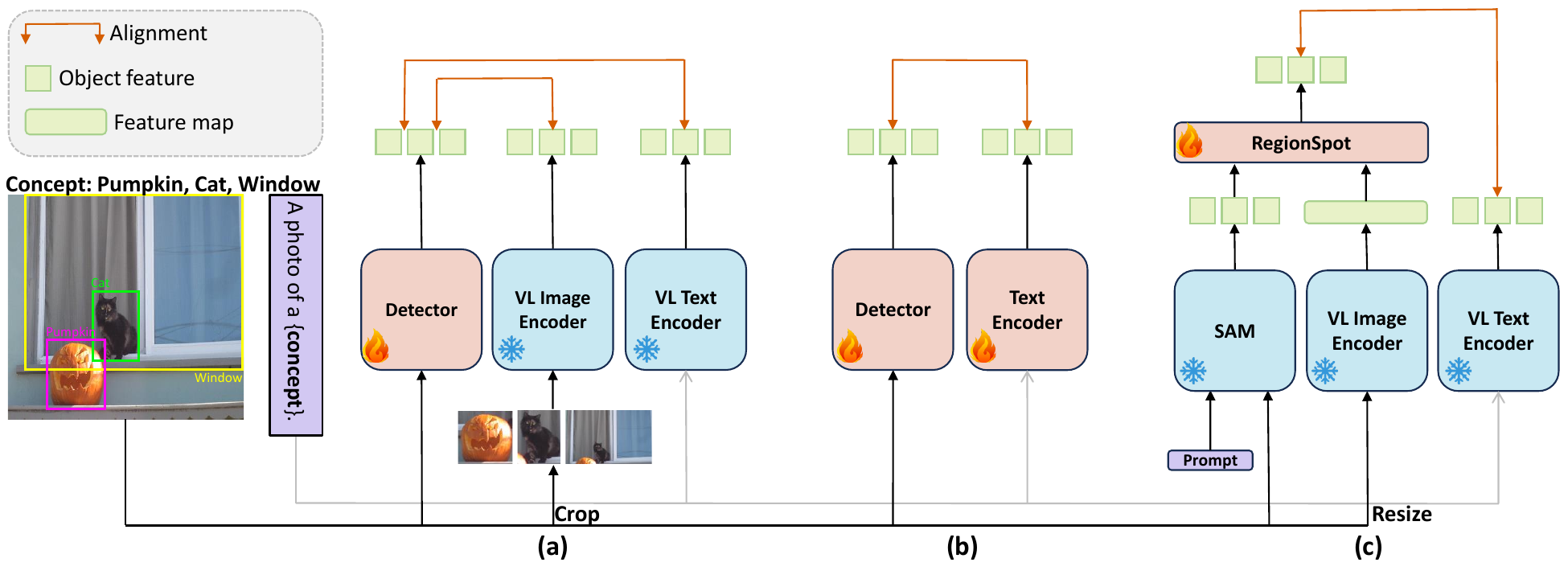}
\vspace{-5pt}
\caption{
\textbf{Illustration of typical region-level visual understanding architecture}.
(a) Learning the region recognition model by distilling image-level ViL representations from cropped regions and incorporating them into a detection model
(\eg, ~\cite{gu2021open}).
(b) Fully fine-tuning both vision and text models with a substantial dataset of region-label pairs.
(c) Our proposed approach integrates pretrained (frozen) localization and ViL models, emphasizing the learning of their representational correlation.
}
\label{fig:tease}
\vspace{-23pt}
\end{figure}

Remarkable progress has been achieved in the realm of purpose-generic image-level Vision-Language (ViL) representation learning, as exemplified by foundation models like CLIP~\cite{radford2021learning} and ALIGN~\cite{jia2021scaling}. These advancements have led to significant performance improvements across a diverse spectrum of vision and multi-modal downstream tasks~\cite{gu2021open, zhou2022conditional}.
The efficacy of these approaches can be {largely attributed} to their utilization of extensive datasets, typically encompassing millions, if not billions, of training samples replete with rich information. In the pursuit of a more nuanced approach to visual analysis, researchers have also ventured into the realm of universal region-level (e.g., objects) comprehension. This is evident in recent research endeavors~\cite{gu2021open,zang2022open,ma2022open, du2022learning,zhong2022regionclip,kuo2022f,lin2022learning,ma2023codet}. 
{A common approach} to this involves learning the semantics of image regions by applying an image-level pre-trained model (e.g., CLIP) to cropped regions, followed by representational distillation using the output of a detection model~\cite{gu2021open, zang2022open}, as depicted in Figure \ref{fig:tease}(a).
However, utilizing individual cropped regions in this design leads to the loss of crucial contextual information, which can hinder recognition performance. 
{~\cite{kuo2022f} introduced an open-world detector with a fixed ViL model, bypassing knowledge distillation. However, the use of ROIAlign~\cite{he2017mask} for region feature extraction poses limitations.}
Furthermore, directly applying an image-level model to isolated local regions is less effective, as the model was pretrained on entire images encompassing both object regions and surrounding context.
An alternative, albeit brute-force, approach revolves around constructing region-level representations from scratch, harnessing an extensive dataset that pairs regions with labels~\cite{li2022grounded,yao2022detclip,zhang2022glipv2,yao2023detclipv2} (Figure \ref{fig:tease}(b)). Nevertheless, this approach grapples with challenges such as the proliferation of noisy pseudo-labels and significant training costs.
Furthermore, significant advancements have materialized in the realm of class-agnostic visual localization techniques, as illustrated by the notable work of SAM~\cite{kirillov2023segment}. This approach is characterized by its distinctive feature—an integration of position-aware localization knowledge, which we consider a valuable complement to the inherent capabilities of ViL models.
Expanding upon this conceptual framework, our research introduces an innovative architectural paradigm at the region level, herein referred to as \textbf{\modelname{}}. This framework seamlessly incorporates large pre-trained ViL and localization models within an efficient training regimen, obviating the necessity for an extensive repository of region-label pairings.
Our methodology centers on the acquisition of the correlation between localization data extracted from `local' regions by the localization model and the semantic representations encompassing the entirety of the image, derived from the ViL model. This strategic approach permits us to circumvent the conventional fine-tuning of both pre-trained models—wherein they remain `frozen' during training—thereby safeguarding the integrity of their rich knowledge and ensuring its maximal utilization, all while mitigating the potential for performance degradation.
To enact this cross-model correlation, we employ the cross-attention mechanism~\cite{vaswani2017attention}. In this configuration, the localization feature assumes the role of the `query', whereas the ViL feature assumes dual roles as both the `key' and `value'. This implementation effectively facilitates the fusion of semantic and localization information in a manner that is amenable to learning and yields substantive efficacy.

Our {\bf contributions} are as follows:
(1) We introduce the concept of integrating off-the-shelf foundation models to tackle region-level visual understanding.
(2) To achieve this objective, we introduce a novel architectural paradigm called \textit{\modelname{}}, which does not necessitate training from scratch. This approach excels in both optimization efficiency and data utilization. By circumventing the fine-tuning of both localization and Vision-Language (ViL) components, our architecture retains its openness and adaptability, welcoming the seamless integration of advancements in both domains.
Extensive experimentation in the context of open-world object understanding confirms the superior performance of our method, even with a substantially smaller number of learnable parameters. Remarkably, \textit{\modelname{}} surpasses the state-of-the-art GLIP-L by 2.9 in mAP, with an even more substantial advantage of 13.1 AP observed for the more intricate rare categories.

\section{Related Work}
\paragraph{Zero-shot in image recognition}

Zero-shot image recognition is the task of recognizing categories that have not been seen during training. In ~\cite{farhadi2009describing} and ~\cite{jayaraman2014zero}, the authors utilized visual attributes to facilitate knowledge transfer to unfamiliar categories. Researchers have also investigated the utilization of class hierarchies, similarities, and object parts to enhance knowledge transfer, as demonstrated in the works of ~\cite{rohrbach2011evaluating, akata2016multi,zhao2017open,xie2020region}.
Recent research has focused on aligning latent image-text embeddings for classifying and describing visual content. ~\cite{frome2013devise} pioneered the establishment of a visual semantic space through deep learning. Subsequently, CLIP~\cite{radford2021learning} and ALIGN~\cite{jia2021scaling} attained impressive results via contrastive learning with extensive collections of image-text pairs, showcasing exceptional performance across diverse benchmarks. In contrast to previous endeavors that primarily addressed image-level recognition, we focus on fine-grained recognition of visual elements at the regional level.

\paragraph{Zero-shot in region understanding}

In zero-shot object recognition, the aim is to enable object detectors to identify categories not encountered during training, such as ~\cite{ren2015faster, sun2021sparse, yan2023universal}. Researchers have explored various methods to bridge the gap between known and unknown categories using pre-trained semantic or textual features ~\cite{socher2013zero,reed2016learning,changpinyo2017predicting}, knowledge graphs ~\cite{salakhutdinov2011learning,wang2018zero}, and more.
Inspired by the zero-shot capabilities of Vision-and-Language (ViL) like CLIP ~\cite{radford2021learning}, several approaches have sought to integrate pretrained Vision-and-Language (ViL) models.
For example, ~\cite{zang2022open, gu2021open,du2022learning} proposed a method to distill learned image embeddings from CLIP for target detection by focusing on cropped proposal regions.
Another approach, RegionCLIP~\cite{zhong2022regionclip} employs a multistage training strategy. 
It starts by generating pseudo-labels from captioning data and then proceeds with region-word contrastive pretraining before transferring the knowledge to the detection task.
~\cite{li2022grounded} took a novel approach by formulating object detection as a grounding problem and incorporating additional grounding data to enhance semantic alignment at both phrase and region levels. Their results demonstrated improved performance, even on fully-supervised detection benchmarks.
~\cite{yao2022detclip} leveraged large-scale image captioning datasets and expanded their knowledge database using generated pseudo-labels, bolstering their detection capabilities.
The use of generated pseudo-labels effectively extended the detectors' generalization ability.

However, these methods face computational challenges and are susceptible to training data inconsistencies and image-level distractions. Differing from these studies, we explore the synergistic benefits of foundation models SAM~\cite{kirillov2023segment} and CLIP~\cite{radford2021learning}. Leveraging their strengths in localization and semantics, we propose an innovative region recognition framework.

\section{Method}
\begin{figure}[]
\centering
\includegraphics[width=0.99\linewidth]{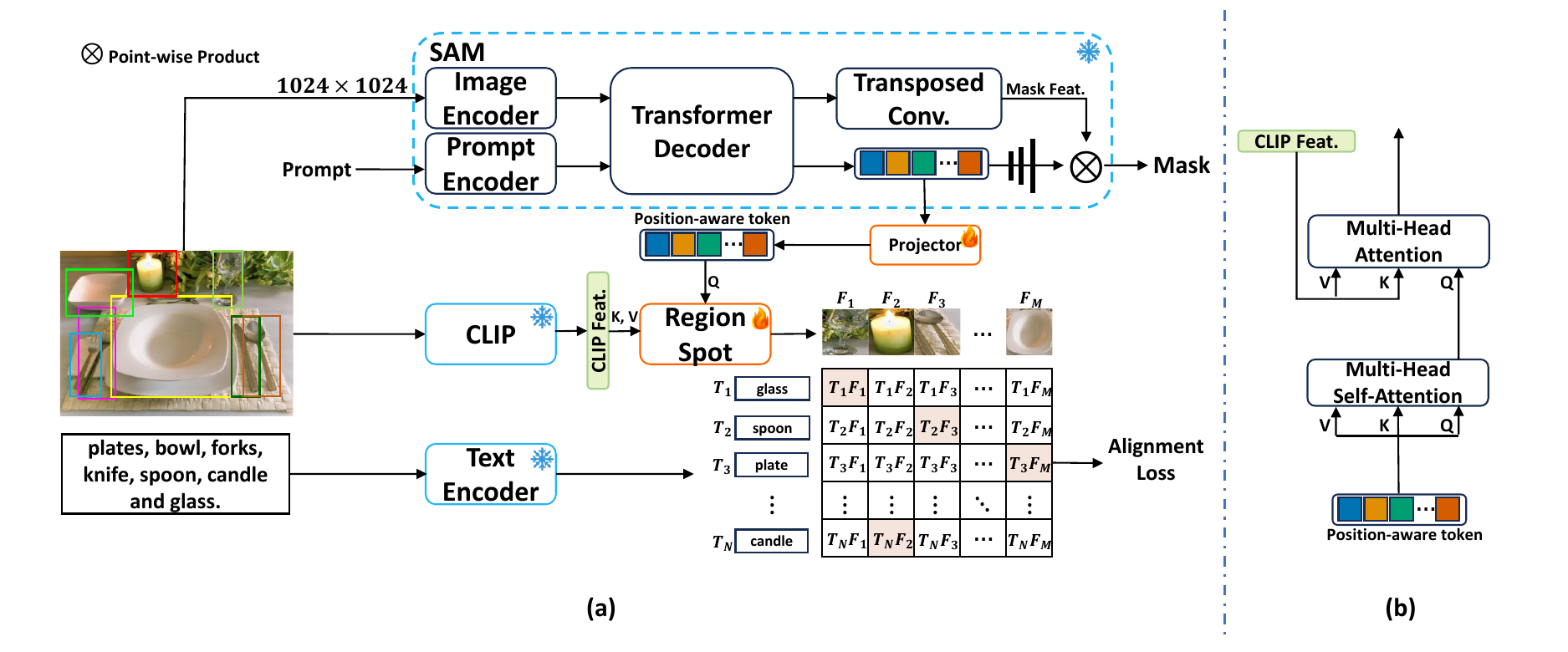}
\vspace{-10pt}
\caption{
Overview of our proposed \modelname{}. (a)
We integrate position-aware tokens from a localization model, such as SAM, with image-level feature maps extracted from a ViL model like CLIP. This integration yields region-level semantic tokens, which are then subjected to region text alignment. (b) Our cross-modal feature interaction design based on the attention mechanism.
}
\label{fig:framework}
\vspace{-12pt}
\end{figure}

Our objective is to employ efficiently a pretrained ViL model and a localization model, trained on extensive data, to achieve region-level representation and understanding. These representations facilitate robust object conceptualization, especially for open-world region recognition. To realize this, as shown in Figure \ref{fig:framework}(a) 
we formulate a new approach, named \modelname. In the following sections, we will begin with a brief introduction to the foundational models in Section \ref{sec:preli}, followed by a comprehensive explanation of our approach with focus on learning region-text alignment across two pretrained models in Section \ref{sec:approch}.

\vspace{-5pt}
\subsection{Foundation Models}
\label{sec:preli}
\textbf{Vision-language foundation models} 
use contrastive learning to map visual and textual data into a shared embedding space through a contrastive loss. This technique, exemplified by CLIP with 400 million text-image pairs ~\cite{radford2021learning}, and ALIGN with 1.8 billion pairs ~\cite{jia2021scaling}, aims to minimize the distances between paired images and texts while maximizing distances between unpaired ones.\\
\textbf{Localization foundation models} have been advanced significantly.
A prominent example is the pioneering SAM model~\cite{kirillov2023segment}, which has been trained on the extensive SA-1B dataset, boasting more than 1 billion automatically generated masks—an unprecedented scale, surpassing existing segmentation datasets by a factor of 400. This dataset also comprises 11 million images. 

SAM comprises three core modules:
(a) Image encoder: Utilizing a ViT-based backbone, this module extracts image features, yielding image embeddings.
(b) Prompt encoder: It encodes positional information from input points, boxes, or masks to facilitate the mask decoder.
(c) Mask decoder: This transformer-based decoder leverages both the extracted image embeddings and prompt tokens to make final mask predictions.
One of SAM's remarkable features is its robust zero-shot generalization to novel data, obviating the need for domain-specific fine-tuning. Thanks to extensive training on a vast repository of prompt-text pairs, SAM demonstrates exceptional proficiency in object localization.

\subsection{Region text alignment with frozen foundation models}
\label{sec:approch}
In this section, we describe how we extract position-aware tokens from the localization foundation model and generate image-level semantic features using the ViL foundation model. We achieve inter-model association through a cross-attention mechanism that facilitates region text alignment.

\paragraph{Region-level position-aware tokens}
In our approach, we utilize manually-annotated object bounding boxes, denoted as $R=\{r_i\}, {i=1,..,N}$, as regions of interest in the images. For each of these regions, represented as $R$, we extract position-aware tokens using the SAM model, denoted as $P=\{p_i\}, {i=1,..,N}$.

As depicted in Figure~\ref{fig:framework}, SAM employs a mask decoder to generate a mask based on a provided prompt. This process utilizes a transformer decoder, similar to the architecture in DETR ~\cite{carion2020end}, to generate an object token. This object token plays a crucial role in predicting the prompt mask, subsequently predicting dynamic MLP weights and performing a point-wise product with the mask features. 
We refer to this resulting token as ``position-aware'' because it encodes essential information about the object, including details about its texture and position. Following this, a projector is applied to map the output dimension of the position-aware token to the image-level feature space as discussed below.

\paragraph{Image-level semantic feature maps}

A single image can encompass multiple objects across numerous categories, capturing integrated context. We can conceptually view an image's feature map as a composition of region embeddings with varying structures. To fully capitalize the ViL model, we resize the input image to the required dimensions without cropping. Subsequently, we input this resized image into the ViL model, yielding the image-level semantic feature map denoted as $V$.

\paragraph{Relating position-aware tokens and semantic feature maps}

Our model, referred to as \modelname{}, efficiently establishes connections between region-level position-aware tokens and image-level semantic feature maps using the cross-attention mechanism ~\cite{vaswani2017attention}. In this mechanism, position-aware tokens $P$ serve as queries, while semantic feature maps $V$ take on the roles of both keys and values. This relationship is formulated as follows:
\begin{equation}
S = \text{Softmax} \left( \frac{F_{p} K_{v}^T}{\sqrt{C}} \right) V_{v},
\end{equation}
where $F_p$ represents a transformation of $P$,  $K_{v}$ and $V_{v}$ are derived from separate linear projections of $V$, and $C$ is the projected feature dimension. This approach, well-established in the literature, has consistently demonstrated its effectiveness in information fusion. In our work, we extend its application to enhance region-level understanding in open-world scenarios.
Specifically, we leverage this mechanism to integrate positional information with semantic content extracted from two distinct models at the regional level, while also considering the broader context from the entire image, as depicted in Figure~\ref{fig:framework}(b).

\paragraph{Loss function}
In line with prior research, we generate text embeddings by processing category texts along with prompt templates, like {\em a photo of {category} in the scene}, using the text encoder. Then, we perform a dot product operation between each semantic token and its corresponding text features to calculate matching scores. These scores can be supervised using the focal loss ~\cite{lin2017focal}.

\paragraph{Zero short inference}
{Following~\cite{zhong2022regionclip}, we
focus on the more challenging region recognition task
by utilizing human-annotated boxes or external region proposal generator.
Inheriting the flexible prompting capability from SAM, 
our model allows for region recognition through prompting.
}

\section{Experiments}
\paragraph{Training data}
In pursuit of a robust training environment, we combined diverse datasets with varying label spaces. Our model's flexible architecture allowed us to seamlessly replace one-hot labels with class name strings.
For training, we utilized publicly available detection datasets, comprising a total of approximately 3 million images. These datasets include Objects 365 (O365) ~\cite{shao2019objects365}, OpenImages (OI) ~\cite{krasin2017openimages}, and V3Det (V3D) ~\cite{wang2023v3det}, each contributing uniquely to the diverse repository.

\begin{itemize}
\item Objects 365 (O365) is a large-scale object detection dataset featuring 365 distinct object categories across 0.66 million images. Our research employs an enriched version with over 10 million bounding boxes, averaging approximately 15.8 object annotations per image.
\item OpenImages (OI) currently stands as the largest public object detection dataset, encompassing about 14.6 million bounding box annotations, equivalent to around 8 annotations per image.
\item V3Det (V3D) distinguishes itself through a hierarchical organization, meticulously structuring up to 13,029 categories within a category tree.
\end{itemize}

\paragraph{Benchmark settings}
In our rigorous evaluation process, we utilized the extensive LVIS detection dataset ~\cite{gupta2019lvis}, which encompasses 1203 categories and 19809 images reserved for validation. 
We do not prioritize the performance on COCO~\cite{lin2014microsoft} which includes only 80 common categories covered by the Objects365 training dataset ~\cite{shao2019objects365}.
This limitation may not adequately assess a model's generalization in an open-world setting.

Since our current emphasis is not on object localization, we utilized ground-truth and class-agnostic bounding boxes from an existing detector to predict categories based on corresponding text descriptions, following the RegionCLIP approach~\cite{zhong2022regionclip}. 
Mean Average Precision (mAP) served as our evaluation metric.

\paragraph{Implementation details}
~\label{sec:implement}
We train \modelname{} using AdamW ~\cite{kingma2014adam} optimizer with the initial learning rate as 2.5 $\times$ $10^{-5}$. 
All models are trained with a mini-batch size 16 on 8 GPUs. The default training schedule is 450K iterations, with the learning rate divided by 10 at 350K and 420K iterations. 
The training process unfolds in two sequential stages: (1) a warm-up phase leveraging the Objects365 to initiate the learning of region-word alignments, and (2) a phase of advanced learning for region-word alignments, utilizing a rich compilation from three diverse object detection datasets.
The model is trained for 450K iterations at each stage.
{
We implement several model variants: 
(1)  \modelnamelite: Integrating the base versions of both SAM and CLIP.
(2) \modelnamepro: Combining the SAM base with the more extensive CLIP large architecture.
(3) \modelnameproplus: 
Further extending \modelnamepro{}
by 
using input image resolution of 336.
}

\subsection{Zero-shot Inference for Region Recognition}

\begin{table}[t]
\setlength\tabcolsep{4pt} 
\centering
\caption{Comparison of open-world zero-shot object recognition performance using ground-truth (GT) boxes,  SAM proposals generate by automatic mask generator, and GLIP boxes on the LVIS dataset. * indicate finetune the CLIP with Adapter. The training time test on one V100 GPU}
\begin{tabular}{l|c|c|c|c|c|c}
\shline
Method & Training Data & Proposals & Times & AP$_{r}$ & AP$_{f}$ & AP$_{all}$ \\
\shline \hline
CLIP-L w/ box & - & GT & - & 40.6 & 59.2 & 48.7 \\
CLIP-L w/ mask & - & GT & - & 40.8 & 59.6 & 49.2 \\
CLIP-L{\scriptsize ↑336} w/ mask & - & GT & - & 43.2 & 59.9 & 49.5\\
CLIP-L{\scriptsize ↑336}* w/ mask & O365, OI, V3D & GT & 0.30k & 46.8 & 63.2 &  53.1\\
\rowcolor{cyan!50}
\modelnamelite & O365, OI, V3D & GT & 0.18k & 42.0 & 65.6 & 53.0 \\
\rowcolor{cyan!50}
\modelnamepro & O365, OI, V3D & GT &  0.18k  & {50.6} & {68.8} & {56.6} \\
\rowcolor{cyan!50}
\modelnameproplus & O365, OI, V3D & GT & 0.20k & \textbf{55.4} & \textbf{68.6} & \textbf{59.9} \\
\hline
CLIP-L{\scriptsize ↑336}* w/ mask & O365, OI, V3D & SAM & 0.30k & 11.3 & 16.4 & 14.5 \\
\rowcolor{cyan!50}
\modelnamepro & O365, OI, V3D & SAM &  0.18k  & {13.1} & {17.3} & {16.1} \\
\rowcolor{cyan!50}
\modelnameproplus & O365, OI, V3D & SAM & 0.20k & \textbf{14.3} & \textbf{19.2} & \textbf{18.2} \\
\hline

GLIP-T (B) & O365 & GLIP-T(B) & 57.5k & 4.2 & 13.6  & 11.3 \\
\rowcolor{cyan!50}
\modelnamelite & O365 & GLIP-T(B) & 0.18k  & 12.7 & 15.7 & 14.1 \\
GLIP-T & O365,GoldG,Cap4M & GLIP-T & 92.1k & 10.1 & 25.5  & 17.2 \\
\rowcolor{cyan!50}
\modelnamelite & O365, OI, V3D & GLIP-T &0.18k  & 20.0 & 24.2 &  21.1 \\
GLIP-L & FourODs,GoldG,Cap24M & GLIP-L & 120k & 17.1 & 35.4 &  26.9 \\
\rowcolor{cyan!50}
\rowcolor{cyan!50}
\modelnameproplus & O365, OI, V3D & GLIP-L & 0.2k & \textbf{30.2} & \textbf{30.0} & \textbf{29.8} \\
\hline
\end{tabular}
\label{tab:main}
\end{table}

\begin{table}[h]
	\centering
	\begin{center}
         \caption{Evaluation of zero-shot object detection on the LVIS minival dataset.} 
        \begin{tabular}{l|c|c|c}
        \shline
       Method & Training Data&AP$_{r}$ & AP  \\
        \shline
        GLIP-L  & FourODs,GoldG,Cap24M & 28.2 & \textbf{37.3} \\
        GroundingDINO-L  &O365,OI,GoldG,Cap4M,COCO,RefC &22.2  & 33.9 \\
        \modelnameproplus  &O365,OI,V3DET &\textbf{33.2} & {36.9} \\
        \hline\end{tabular}
	
		\label{tab:minidet}
		\end{center}
        \vspace{-10pt}
\end{table}

\paragraph{Zero-shot object detection on LVIS Val v1.0} 
Results on the LVIS benchmark are presented in Table~\ref{tab:main}. With ground-truth bounding boxes as region proposals, our model substantially surpasses the CLIP baselines (which applies CLIP on image crops) by a large margin (\eg, \textbf{48.7, 49.2} \vs \textbf{59.9}). 
For fair comparison, we finetune the CLIP withe adapter, 
our model substantially surpasses the CLIP by a large margin.
Moreover, in simulation of real-world cases, we move forward to test our method with noisy region proposals generated from off-the-shelf proposal generator.
We first employ SAM as a proposal generator, inputting dense grid points to automatically generate proposals.
It can be seen that RegionSpot still consistently outperforms CLIP (\eg, \textbf{14.5} \vs \textbf{18.2} on AP$_{all}$) in this case, demonstrating the robustness of our method. 

To fully exploit the potential of our method and synergy with the advancements of open world object detection (OWD), we further utilize region proposals from state-of-the-art OWD models, \ie, GLIP, as our region prompts. Comparing with GLIP-T trained solely on the objects365 dataset, we can observe a considerable performance gain achieved by RegionSpot (\eg, \textbf{4.2} \vs \textbf{12.7} on AP$_{r}$ and \textbf{11.3} AP \vs \textbf{14.1} on AP$_{all}$). After scaling up the training data and use 336 resolution image as input, our models maintains superior performances over their GLIP counterparts. For instance, \modelnameproplus surpasses GLIP-T by \textbf{17.6} AP$_{r}$ with less training data, showing compelling scaling behavior with data scale and input resolution. 
{For more extensive evaluation, we also utilize bounding boxes generated by GLIP-L as prompts. It is noteworthy that RegionSpot achieves an impressive \textbf{13.1} increase in AP$_{\textbf{r}}$ compared to GLIP-L, even when trained on less data at higher efficiency.} 
Despite using a noisy box, we were still able to achieve promising results, thanks to the robust localization ability of SAM. Additional experiments, including the ViLD protocol, can be found in the Appendix
\vspace{-5pt}
\paragraph{Open vocabulary object detection under ViLD-protocal}
To thoroughly evaluate our method, we conducted experiments using the ViLD protocol \cite{gu2021open}, training on base categories and testing on novel ones with the LVIS AP metric. 
For fair comparsion, all the method training only use the LVIS-base dataset and use the RPN  from RegionCLIP as proposal generator.
We  also adapted our training to the LVIS-base dataset. 
As shown in Table ~\ref{tab:vild_setting}, RegionSpot demonstrates competitive performance. It outperforms the similarly frozen-backbone F-VLM by 1.1 AP$_{r}$ .
When we compared to RegionCLIP, which benefits from additional caption pretraining, RegionSpot significantly outperforms the pretrained version of RegionCLIP by 2.6 when utilizing same RPN.
\begin{table}[h]
    \centering
     \caption{Comparison under the ViLD protocol~\cite{gu2021open}. All methods use the ResNet50 backbone. * indicate  pre-training with CC-3M}
    \begin{tabular}{l|c|c|c|c}
    \hline
    Method & Proposals &Trainable
Backbone & AP$_{\text{r}}$ &AP$_{\text{all}}$  \\
    \hline
    ViLD & RPN & \Checkmark & 16.1 & \textit{\textcolor{gray}{22.5}} \\
    RegionCLIP* &RPN&\Checkmark & 17.1 & \textit{\textcolor{gray}{28.2}} \\
    Detic-ViLD & RPN &\Checkmark & 17.8 & \textit{\textcolor{gray}{ 26.8}}  \\
    F-VLM & RPN &\XSolidBrush&  18.6  & \textit{\textcolor{gray}{ 24.2}}  \\
    RegionSpot  & RPN & \XSolidBrush & \textbf{19.7}  & \textit{\textcolor{gray}{ 25.0}} \\
    \hline
    \end{tabular}
    \label{tab:vild_setting}
    \vspace{-5pt}
\end{table}

\vspace{-5pt}
{\paragraph{Zero-shot object detection on LVIS minival5k~\cite{kamath2021mdetr}}
To fully exploit the potential of our method, we report on MiniVal containing 5,000 images introduced in MDETR~\cite{kamath2021mdetr}.
We use the output proposals from GLIP as the prompt.
As shown in Table ~\ref{tab:minidet}, although we use 9x less training data, our model maintains superior performances over GLIP-L by 5.0 on APr. 
Further, our method also surpasses Grounding DINO-L (which even uses a more advanced detector) by 11.0 in APr.
\begin{table}[]
	\centering
	\begin{center}
        \caption{Evaluation of zero-shot instance segmentation on the LVIS minival dataset.} 
        \begin{tabular}{l | c|c|c|c}
        \shline
       Method & AP$_{r}$ & AP$_{c}$ & AP$_{f}$ & AP  \\
        \shline
        X-Decoder  & - & - & - & 9.4 \\
        OpenSeed    & - & - & -  & 19.6 \\
        \modelnameproplus  & \textbf{21.5} & \textbf{25.0} & \textbf{23.2} & \textbf{23.5} \\
        \hline\end{tabular}	
		\label{tab:mask}
		\end{center}
        \vspace{-14pt}
\end{table}
\paragraph{Zero-shot instance segmentation}
We evaluate the performance of instance segmentation in a zero-shot setting using the LVIS dataset~\cite{gupta2019lvis}. By leveraging the output from GLIP as the prompt, we direct it to \modelname{} for mask and class prediction. 
The mask AP is evaluated using the released X-Decoder~\cite{zou2023generalized} and OpenSeeD~\cite{zhang2023simple}, both of which are trained with mask-text pairs. 
Impressively, as indicated in Table~\ref{tab:mask}, \modelname{} outstrips X-Decoder and OpenSeeD by margins of 14.1  and 3.9 in AP, respectively. 
These outcomes suggest that our proposed \modelname{} can effectively harness the foundational model's capabilities to achieve more accurate region understanding.
\vspace{-5pt}
\paragraph{Zero-shot object detection on ODinW~\cite{li2022elevater}}
This benchmark was designed to evaluate model performance in real-world scenarios. To accurately evaluate recognition capabilities, we filter the dataset to include only those with more than three categories. The AP for each dataset is reported in Table ~\ref{tab:odinw}. Impressively, \modelnameproplus{}, utilizing GLIP-L proposals, surpasses GLIP-L by a margin of 2.5 AP, attributable to its precise region recognition.  Furthermore, our method exceeds the performance of GroundingDINO, even though it employs a more advanced detector.
}
\begin{table}[]
\setlength\tabcolsep{4pt} 

	\centering
 \vspace{-10pt}
	\begin{center}
        \caption{Evaluation of zero-shot object detection on the ODinW dataset.} 
        \begin{tabular}{l|c|c|c|c|c|c}
        \shline
       Method & Aerial. Drone. & Aquarium & PascalVOC& shellfish& vehicles& Avg. \\
        \shline
        
        GroundingDINO-T  & 10.3  & 17.5 & 55.7 & 29.5 & 58.5 & 34.3 \\
        GLIP-T  & 12.5  & 18.4 & 56.2 & 26.3 & 56.0 & 33.8 \\
        GLIP-L & 7.1  & 26.9 & 61.7 & 68.9 & 57.3 & 44.4 \\
        \modelnamelite  & 13.1  & 20.1 & 58.2 & 30.1 & 57.2 & 35.7 \\
        \modelnameproplus  & \textbf{14.2}  & \textbf{27.2} & \textbf{62.7} & \textbf{69.3} & \textbf{61.3} & \textbf{46.9} \\
        \hline\end{tabular}
		\label{tab:odinw}
		\end{center}
        \vspace{-18pt}
\end{table}

\vspace{-5pt}
\subsection{Ablation Study}
\vspace{-5pt}
We conducted an ablation study for \modelname{}-\texttt{BL} using the boxes generated by GLIP. Unless otherwise mentioned, training was performed on three different detection datasets.
\begin{table}[h]
\centering
\caption{\textbf{Ablation experiments on LVIS.}
{\,\textbf{(a)} The effective of CLIP vision encoder};
{\,\textbf{(b)} Position-aware tokens selection };
{\,\textbf{(c)} Depth of \modelname{}.} 
}
\subfloat[]
{
    \begin{minipage}{0.3\linewidth}{
    \begin{center}
    \begin{tabular}{l| c }
    \shline
    CLIP  &  AP  \\
    \shline
    w/o CLIP vision. & 8.0 \\
    + CLIP feat. map & 22.1  \\
    + Class token & \textbf{23.7} \\
    \hline\end{tabular}
    \end{center}
    }
    \label{tab:clip}
    \end{minipage}
}
\subfloat[]
{
    \begin{minipage}{0.3\linewidth}{
    \begin{center}
    \begin{tabular}{l| c}
    \shline
    Position-aware tokens &  AP   \\
    \shline
    Prompt encoder  & 18.6  \\
    Transformer & \textbf{23.7} \\
    MLP & 20.4  \\
    \hline\end{tabular}
    \end{center}}
    \label{tab:sam}
    \end{minipage}
}
\subfloat[]
{
    \begin{minipage}{0.3\linewidth}{
    \begin{center}
    \begin{tabular}{c | c}
    \shline
    Depth &  AP \\
    \shline
    1  & 23.2  \\
    3 & \textbf{23.7}\\
    6  & 22.8 \\
    \hline\end{tabular}
    \end{center}}
    \label{tab:layers}
    \end{minipage}
}
\vspace{-20pt}
\end{table}
\vspace{-5pt}
\paragraph{Enhancement with CLIP vision embedding}
We conjugate that a key ingredient with \modelname{} is the use of semantic information from CLIP vision encoder.  To validate this assertion, we began our evaluation without the CLIP feature and subsequently integrated the class token output from the CLIP vision encoder. Results in Table~\ref{tab:clip} demonstrate that: 
(1) The CLIP feature offers a significant boost, pushing the baseline without CLIP vision encoder to \textbf{22.1}, which suggests inherent semantic limitations in SAM.
(2) More notably, the inclusion of CLIP enhances overall performance to \textbf{23.7}. This underscores the potential of the class token to encapsulate global information from the entire image.

\vspace{-5pt}
\paragraph{Position-aware tokens selection in SAM}
The position-aware tokens are generated by intermediary module in the SAM.
We examined various locations for this generation, specifically after the Prompt encoder, the Transformer decoder, and the MLP within the SAM. 
Results presented in Table~\ref{tab:sam} indicate that generating output tokens after the Transformer decoder yields the best performance.
This observation is expected since tokens derived from the Prompt encoder are relatively undeveloped. 
Surprisingly, it can outperform GLIP (\textit{i.e.}, \textbf{18.6} vs. \textbf{17.2}).
Moreover, there is a performance decline after the MLP, which can be attributed to dimensional reduction.

\paragraph{Module architecture}
Another pivotal aspect of \modelname{} is the depth of model. To assess its impact, we experimented by varying the depth of our model. As indicated in Table~\ref{tab:layers}, it is imperative for the model to have a sufficiently large depth, such as 3 blocks, without being excessive.
\vspace{-5pt}
\paragraph{Prompt engineering}
\begin{table}[]
\centering
\vspace{-10pt}
\caption{\textbf{Ablation experiments on LVIS.}
{\,\textbf{(a)} The effective of  propmpt engineering};
{\,\textbf{(b)} The effective of  SAM  }
}
\subfloat[]
{
    \begin{minipage}{0.45\linewidth}{
    \begin{center}
    \begin{tabular}{l | c | c | c}
    \shline
       Prompt & AP$_{r}$ & AP$_{f}$ & AP  \\
        \shline
        baseline &  19.6  & 21.2 & 18.5 \\
        w/ mutiple boxes prompt &  23.2  & 25.0 & 22.1  \\
        w/ text prompt &  \textbf{24.9} & \textbf{25.5} & \textbf{23.7}   \\
    \hline
    \end{tabular}
    \end{center}
    }
    \label{tab:trick}
    \end{minipage}
}
\subfloat[]
{
    \begin{minipage}{0.45\linewidth}{
    \begin{center}
    \begin{tabular}{l| c | c}
    \shline
    SAM &  box AP$_{r}$ & mask AP$_{r}$   \\
    \shline
    ViT-B  & 24.9 & 22.8  \\
    ViT-L  & 24.7 & 23.6 \\
    \hline\end{tabular}
    \end{center}}
    \label{tab:sam}
    \end{minipage}
}
\vspace{-15pt}
\end{table}
Finally, we carried out an ablation study focusing on prompt engineering, incorporating both box prompts in SAM and text prompts in the text encoder. As evidenced by the results in Table~\ref{tab:trick}:
(1) Leveraging multiple boxes as prompts in SAM boosts performance, achieving an AP of 22.1. This enhancement is credited to the self-attention mechanism of \modelname{}, which adeptly fuses information from varied regions.
(2) Further utilizing text prompts results in a modest performance boost, specifically an increase of 1.6 AP.
\vspace{-5pt}
\paragraph{Ablation study of SAM model}
We conjugate that a key ingredient is the position-aware information from SAM.
We evaluating the impact of different SAM model sizes, such as ViT-L, is essential.
We conducted  experiments with varying SAM model sizes.
As shown in the Table~\ref{tab:sam}, our findings are summarized as follows:
(1) Impact of SAM Model Size: Our results indicate that the use of larger SAM models (e.g., SAM-L) improves mask AP due to the higher quality of mask generation. However, for box AP, the improvement is not significant. This is because the SAM mask token primarily contributes position-aware knowledge, which is already sufficiently captured by ViT-B and ViT-L.
(2) Choice of SAM Model: Given our focus on region recognition, we opted for SAM-B, balancing performance and computational efficiency.

\begin{figure}[h]
   \includegraphics[width=0.97\linewidth]{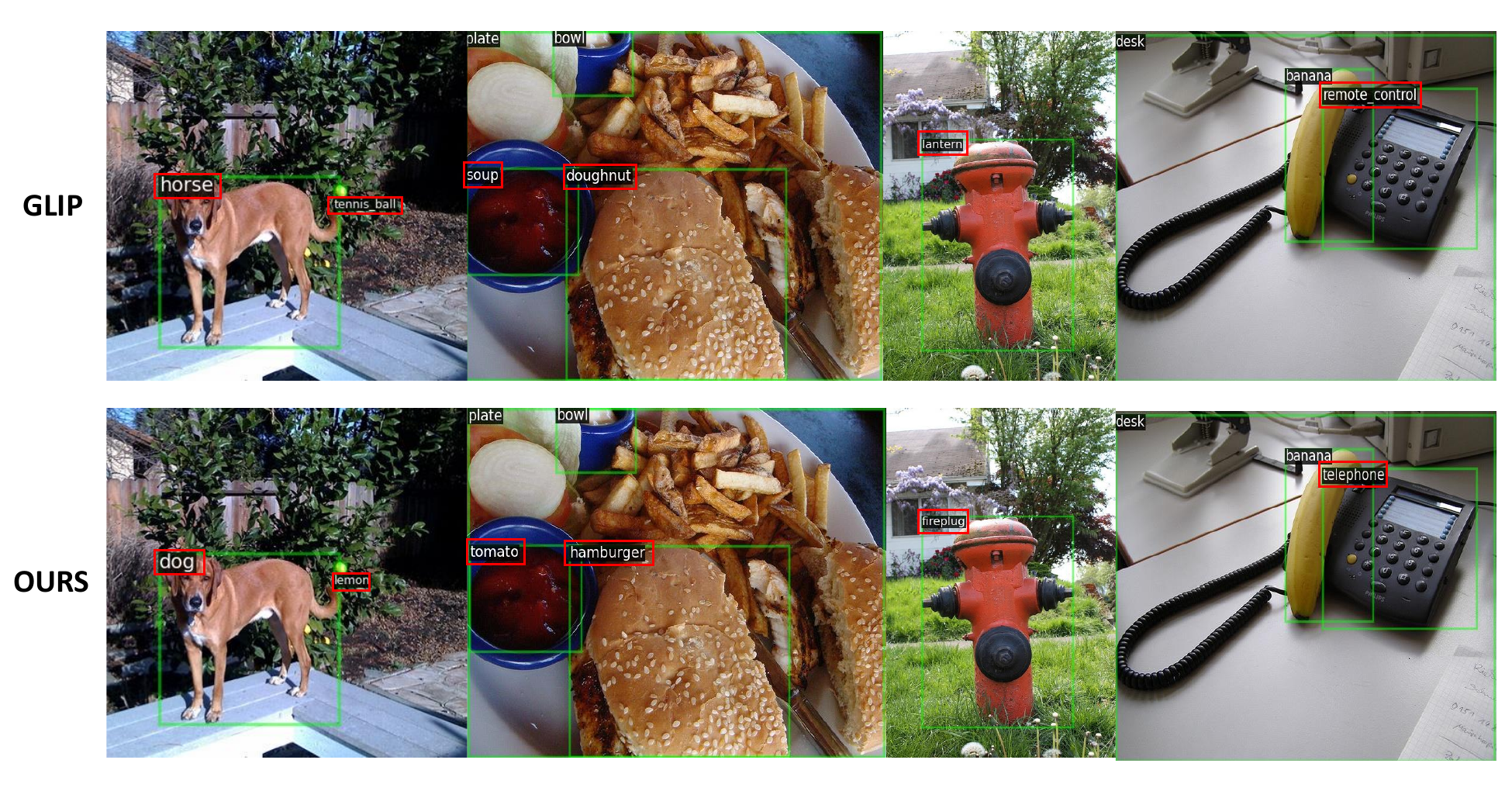}
   \caption{
Qualitative prediction results of GLIP-T~\cite{li2022grounded} (first row) and \modelname{} (second row) on the LVIS dataset~\cite{gupta2019lvis}. Our model recognizes the objects more accurately. 
Best viewed when zooming-in.
}
\label{fig:recognition}
\vspace{-10pt}
\end{figure} 
\vspace{-5pt}
\subsection{Visualization}
\begin{figure*}[]
    \centering
\includegraphics[width=0.97\linewidth]{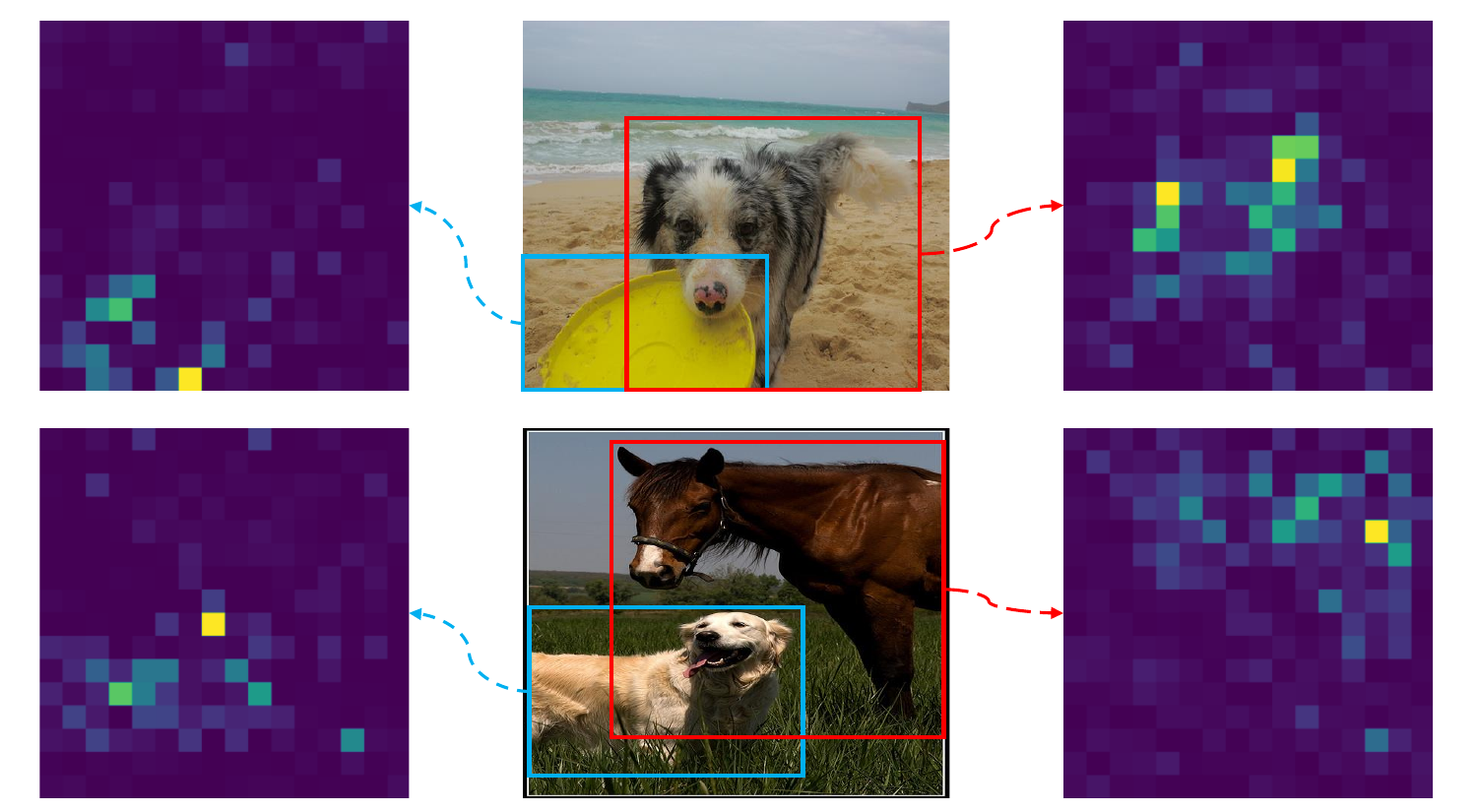}
\caption{Cross-attention maps in \modelname{}.
These maps show that the position-aware token aligns effectively with the semantic feature map of the entire image.
In each row, the blue and red boxes are corresponding to the left and right maps respectively.
}
\label{fig:attn_map}
\end{figure*}
\vspace{-5pt}
\paragraph{Result visualization}
In Figure~\ref{fig:recognition}, we present the results of bounding region recognition on the LVIS ~\cite{gupta2019lvis} dataset, comparing between GLIP and \modelname{}.
To assess the zero-shot recognition capability, we employ the same bounding boxes for both models. As observed, \modelname{} can distinguish even subtle differences, recognizing smaller objects like ``lemon'' and ``tennis ball'' and similar objects like ``lantern'' and ``fireplug''.
Notably, \modelname{} stands out in terms of the accuracy of its label predictions, especially within the category of rare classes.
\vspace{-5pt}
\paragraph{Model behavior visualization}
To gain more intuitive understanding on the effect  brought by our \modelname{}, we examine the cross attention map on LVIS~\cite{gupta2019lvis}. 
We take the output tokens as `query' and CLIP feature map as `key' and `value'.
For clearer visualization, we omit the class token from the CLIP semantic feature. The resulting attention map clearly depicts the correspondence between the position-aware tokens generated by SAM and the feature map produced by CLIP.
Using this arrangement, we gain a visual insight into how \modelname{} establishes connections between distinct features. As depicted in Figure~\ref{fig:attn_map}, the attention maps vividly showcase \modelname{} capability to seamlessly incorporate both SAM and CLIP. Such visualizations serve a dual purpose: they highlight the efficacy of our method, and simultaneously, shed light on the intricate mechanisms underpinning the \modelname{}.
\vspace{-5pt}
\section{Conclusions and limitations}
\vspace{-5pt}
\label{sec:limi_con}
In this study, we introduce \modelname{}, a novel and efficient framework leveraging frozen vision and vision-language foundation models for region recognition, eliminating the need for training from scratch.
To fully exploit knowledge in pretrained models and minimize the training overhead, we keep both foundation models frozen and focus optimization efforts solely on a lightweight attention-based knowledge integration module.
Extensive experiments in the context of open-world object understanding confirms the superior performance of our method, even with a substantially smaller number of learnable parameters, which distinguishes our method and enables efficient training.
Impressively, \textit{\modelname{}} outperforms the leading GLIP-L by 2.9 in mAP, and this lead grows to 13.1 when considering complex rare categories.
While our method advances open world region understanding, it still not unleash potential capabilities from the foundmental models,  such as the automatic localization ability from SAM, which could reduce reliance on external region proposal mechanisms for object detection and enhance versatility. We leave this for further investigation.

\bibliographystyle{main}
\bibliography{main}

\begin{thebibliography}{10}\itemsep=-1pt

\bibitem{akata2016multi}
Z.~Akata, M.~Malinowski, M.~Fritz, and B.~Schiele.
\newblock Multi-cue zero-shot learning with strong supervision.
\newblock In {\em Proceedings of the IEEE Conference on Computer Vision and Pattern Recognition}, pages 59--68, 2016.

\bibitem{carion2020end}
N.~Carion, F.~Massa, G.~Synnaeve, N.~Usunier, A.~Kirillov, and S.~Zagoruyko.
\newblock End-to-end object detection with transformers.
\newblock In {\em European conference on computer vision}, pages 213--229. Springer, 2020.

\bibitem{changpinyo2017predicting}
S.~Changpinyo, W.-L. Chao, and F.~Sha.
\newblock Predicting visual exemplars of unseen classes for zero-shot learning.
\newblock In {\em Proceedings of the IEEE international conference on computer vision}, pages 3476--3485, 2017.

\bibitem{du2022learning}
Y.~Du, F.~Wei, Z.~Zhang, M.~Shi, Y.~Gao, and G.~Li.
\newblock Learning to prompt for open-vocabulary object detection with vision-language model.
\newblock In {\em Proceedings of the IEEE/CVF Conference on Computer Vision and Pattern Recognition}, pages 14084--14093, 2022.

\bibitem{farhadi2009describing}
A.~Farhadi, I.~Endres, D.~Hoiem, and D.~Forsyth.
\newblock Describing objects by their attributes.
\newblock In {\em 2009 IEEE conference on computer vision and pattern recognition}, pages 1778--1785. IEEE, 2009.

\bibitem{frome2013devise}
A.~Frome, G.~S. Corrado, J.~Shlens, S.~Bengio, J.~Dean, M.~Ranzato, and T.~Mikolov.
\newblock Devise: A deep visual-semantic embedding model.
\newblock {\em Advances in neural information processing systems}, 26, 2013.

\bibitem{gu2021open}
X.~Gu, T.-Y. Lin, W.~Kuo, and Y.~Cui.
\newblock Open-vocabulary object detection via vision and language knowledge distillation.
\newblock {\em arXiv preprint arXiv:2104.13921}, 2021.

\bibitem{gupta2019lvis}
A.~Gupta, P.~Dollar, and R.~Girshick.
\newblock Lvis: A dataset for large vocabulary instance segmentation.
\newblock In {\em Proceedings of the IEEE/CVF conference on computer vision and pattern recognition}, pages 5356--5364, 2019.

\bibitem{he2017mask}
K.~He, G.~Gkioxari, P.~Doll{\'a}r, and R.~Girshick.
\newblock Mask r-cnn.
\newblock In {\em Proceedings of the IEEE international conference on computer vision}, pages 2961--2969, 2017.

\bibitem{jayaraman2014zero}
D.~Jayaraman and K.~Grauman.
\newblock Zero-shot recognition with unreliable attributes.
\newblock {\em Advances in neural information processing systems}, 27, 2014.

\bibitem{jia2021scaling}
C.~Jia, Y.~Yang, Y.~Xia, Y.-T. Chen, Z.~Parekh, H.~Pham, Q.~Le, Y.-H. Sung, Z.~Li, and T.~Duerig.
\newblock Scaling up visual and vision-language representation learning with noisy text supervision.
\newblock In {\em International conference on machine learning}, pages 4904--4916. PMLR, 2021.

\bibitem{kamath2021mdetr}
A.~Kamath, M.~Singh, Y.~LeCun, G.~Synnaeve, I.~Misra, and N.~Carion.
\newblock Mdetr-modulated detection for end-to-end multi-modal understanding.
\newblock In {\em Proceedings of the IEEE/CVF International Conference on Computer Vision}, pages 1780--1790, 2021.

\bibitem{kingma2014adam}
D.~P. Kingma and J.~Ba.
\newblock Adam: A method for stochastic optimization.
\newblock {\em arXiv preprint arXiv:1412.6980}, 2014.

\bibitem{kirillov2023segment}
A.~Kirillov, E.~Mintun, N.~Ravi, H.~Mao, C.~Rolland, L.~Gustafson, T.~Xiao, S.~Whitehead, A.~C. Berg, W.-Y. Lo, et~al.
\newblock Segment anything.
\newblock {\em arXiv preprint arXiv:2304.02643}, 2023.

\bibitem{krasin2017openimages}
I.~Krasin, T.~Duerig, N.~Alldrin, V.~Ferrari, S.~Abu-El-Haija, A.~Kuznetsova, H.~Rom, J.~Uijlings, S.~Popov, A.~Veit, et~al.
\newblock Openimages: A public dataset for large-scale multi-label and multi-class image classification.
\newblock {\em Dataset available from https://github. com/openimages}, 2(3):18, 2017.

\bibitem{kuo2022f}
W.~Kuo, Y.~Cui, X.~Gu, A.~Piergiovanni, and A.~Angelova.
\newblock F-vlm: Open-vocabulary object detection upon frozen vision and language models.
\newblock {\em arXiv preprint arXiv:2209.15639}, 2022.

\bibitem{li2022elevater}
C.~Li, H.~Liu, L.~Li, P.~Zhang, J.~Aneja, J.~Yang, P.~Jin, H.~Hu, Z.~Liu, Y.~J. Lee, et~al.
\newblock Elevater: A benchmark and toolkit for evaluating language-augmented visual models.
\newblock {\em Advances in Neural Information Processing Systems}, 35:9287--9301, 2022.

\bibitem{li2022grounded}
L.~H. Li, P.~Zhang, H.~Zhang, J.~Yang, C.~Li, Y.~Zhong, L.~Wang, L.~Yuan, L.~Zhang, J.-N. Hwang, et~al.
\newblock Grounded language-image pre-training.
\newblock In {\em Proceedings of the IEEE/CVF Conference on Computer Vision and Pattern Recognition}, pages 10965--10975, 2022.

\bibitem{lin2022learning}
C.~Lin, P.~Sun, Y.~Jiang, P.~Luo, L.~Qu, G.~Haffari, Z.~Yuan, and J.~Cai.
\newblock Learning object-language alignments for open-vocabulary object detection.
\newblock {\em arXiv preprint arXiv:2211.14843}, 2022.

\bibitem{lin2017focal}
T.-Y. Lin, P.~Goyal, R.~Girshick, K.~He, and P.~Doll{\'a}r.
\newblock Focal loss for dense object detection.
\newblock In {\em Proceedings of the IEEE international conference on computer vision}, pages 2980--2988, 2017.

\bibitem{lin2014microsoft}
T.-Y. Lin, M.~Maire, S.~Belongie, J.~Hays, P.~Perona, D.~Ramanan, P.~Doll{\'a}r, and C.~L. Zitnick.
\newblock Microsoft coco: Common objects in context.
\newblock In {\em Computer Vision--ECCV 2014: 13th European Conference, Zurich, Switzerland, September 6-12, 2014, Proceedings, Part V 13}, pages 740--755. Springer, 2014.

\bibitem{ma2023codet}
C.~Ma, Y.~Jiang, X.~Wen, Z.~Yuan, and X.~Qi.
\newblock Codet: Co-occurrence guided region-word alignment for open-vocabulary object detection.
\newblock {\em arXiv preprint arXiv:2310.16667}, 2023.

\bibitem{ma2022open}
Z.~Ma, G.~Luo, J.~Gao, L.~Li, Y.~Chen, S.~Wang, C.~Zhang, and W.~Hu.
\newblock Open-vocabulary one-stage detection with hierarchical visual-language knowledge distillation.
\newblock In {\em Proceedings of the IEEE/CVF Conference on Computer Vision and Pattern Recognition}, pages 14074--14083, 2022.

\bibitem{radford2021learning}
A.~Radford, J.~W. Kim, C.~Hallacy, A.~Ramesh, G.~Goh, S.~Agarwal, G.~Sastry, A.~Askell, P.~Mishkin, J.~Clark, et~al.
\newblock Learning transferable visual models from natural language supervision.
\newblock In {\em International conference on machine learning}, pages 8748--8763. PMLR, 2021.

\bibitem{reed2016learning}
S.~Reed, Z.~Akata, H.~Lee, and B.~Schiele.
\newblock Learning deep representations of fine-grained visual descriptions.
\newblock In {\em Proceedings of the IEEE conference on computer vision and pattern recognition}, pages 49--58, 2016.

\bibitem{ren2015faster}
S.~Ren, K.~He, R.~Girshick, and J.~Sun.
\newblock Faster r-cnn: Towards real-time object detection with region proposal networks.
\newblock {\em Advances in neural information processing systems}, 28, 2015.

\bibitem{rohrbach2011evaluating}
M.~Rohrbach, M.~Stark, and B.~Schiele.
\newblock Evaluating knowledge transfer and zero-shot learning in a large-scale setting.
\newblock In {\em CVPR 2011}, pages 1641--1648. IEEE, 2011.

\bibitem{salakhutdinov2011learning}
R.~Salakhutdinov, A.~Torralba, and J.~Tenenbaum.
\newblock Learning to share visual appearance for multiclass object detection.
\newblock In {\em CVPR 2011}, pages 1481--1488. IEEE, 2011.

\bibitem{shao2019objects365}
S.~Shao, Z.~Li, T.~Zhang, C.~Peng, G.~Yu, X.~Zhang, J.~Li, and J.~Sun.
\newblock Objects365: A large-scale, high-quality dataset for object detection.
\newblock In {\em Proceedings of the IEEE/CVF international conference on computer vision}, pages 8430--8439, 2019.

\bibitem{socher2013zero}
R.~Socher, M.~Ganjoo, C.~D. Manning, and A.~Ng.
\newblock Zero-shot learning through cross-modal transfer.
\newblock {\em Advances in neural information processing systems}, 26, 2013.

\bibitem{sun2021sparse}
P.~Sun, R.~Zhang, Y.~Jiang, T.~Kong, C.~Xu, W.~Zhan, M.~Tomizuka, L.~Li, Z.~Yuan, C.~Wang, et~al.
\newblock Sparse r-cnn: End-to-end object detection with learnable proposals.
\newblock In {\em Proceedings of the IEEE/CVF conference on computer vision and pattern recognition}, pages 14454--14463, 2021.

\bibitem{vaswani2017attention}
A.~Vaswani, N.~Shazeer, N.~Parmar, J.~Uszkoreit, L.~Jones, A.~N. Gomez, {\L}.~Kaiser, and I.~Polosukhin.
\newblock Attention is all you need.
\newblock {\em Advances in neural information processing systems}, 30, 2017.

\bibitem{wang2023v3det}
J.~Wang, P.~Zhang, T.~Chu, Y.~Cao, Y.~Zhou, T.~Wu, B.~Wang, C.~He, and D.~Lin.
\newblock V3det: Vast vocabulary visual detection dataset.
\newblock {\em arXiv preprint arXiv:2304.03752}, 2023.

\bibitem{wang2018zero}
X.~Wang, Y.~Ye, and A.~Gupta.
\newblock Zero-shot recognition via semantic embeddings and knowledge graphs.
\newblock In {\em Proceedings of the IEEE conference on computer vision and pattern recognition}, pages 6857--6866, 2018.

\bibitem{xie2020region}
G.-S. Xie, L.~Liu, F.~Zhu, F.~Zhao, Z.~Zhang, Y.~Yao, J.~Qin, and L.~Shao.
\newblock Region graph embedding network for zero-shot learning.
\newblock In {\em Computer Vision--ECCV 2020: 16th European Conference, Glasgow, UK, August 23--28, 2020, Proceedings, Part IV 16}, pages 562--580. Springer, 2020.

\bibitem{yan2023universal}
B.~Yan, Y.~Jiang, J.~Wu, D.~Wang, P.~Luo, Z.~Yuan, and H.~Lu.
\newblock Universal instance perception as object discovery and retrieval.
\newblock In {\em Proceedings of the IEEE/CVF Conference on Computer Vision and Pattern Recognition}, pages 15325--15336, 2023.

\bibitem{yao2023detclipv2}
L.~Yao, J.~Han, X.~Liang, D.~Xu, W.~Zhang, Z.~Li, and H.~Xu.
\newblock Detclipv2: Scalable open-vocabulary object detection pre-training via word-region alignment.
\newblock In {\em Proceedings of the IEEE/CVF Conference on Computer Vision and Pattern Recognition}, pages 23497--23506, 2023.

\bibitem{yao2022detclip}
L.~Yao, J.~Han, Y.~Wen, X.~Liang, D.~Xu, W.~Zhang, Z.~Li, C.~Xu, and H.~Xu.
\newblock Detclip: Dictionary-enriched visual-concept paralleled pre-training for open-world detection.
\newblock {\em Advances in Neural Information Processing Systems}, 35:9125--9138, 2022.

\bibitem{zang2022open}
Y.~Zang, W.~Li, K.~Zhou, C.~Huang, and C.~C. Loy.
\newblock Open-vocabulary detr with conditional matching.
\newblock In {\em European Conference on Computer Vision}, pages 106--122. Springer, 2022.

\bibitem{zhang2023simple}
H.~Zhang, F.~Li, X.~Zou, S.~Liu, C.~Li, J.~Gao, J.~Yang, and L.~Zhang.
\newblock A simple framework for open-vocabulary segmentation and detection.
\newblock {\em arXiv preprint arXiv:2303.08131}, 2023.

\bibitem{zhang2022glipv2}
H.~Zhang, P.~Zhang, X.~Hu, Y.-C. Chen, L.~Li, X.~Dai, L.~Wang, L.~Yuan, J.-N. Hwang, and J.~Gao.
\newblock Glipv2: Unifying localization and vision-language understanding.
\newblock {\em Advances in Neural Information Processing Systems}, 35:36067--36080, 2022.

\bibitem{zhao2017open}
H.~Zhao, X.~Puig, B.~Zhou, S.~Fidler, and A.~Torralba.
\newblock Open vocabulary scene parsing.
\newblock In {\em Proceedings of the IEEE International Conference on Computer Vision}, pages 2002--2010, 2017.

\bibitem{zhong2022regionclip}
Y.~Zhong, J.~Yang, P.~Zhang, C.~Li, N.~Codella, L.~H. Li, L.~Zhou, X.~Dai, L.~Yuan, Y.~Li, et~al.
\newblock Regionclip: Region-based language-image pretraining.
\newblock In {\em Proceedings of the IEEE/CVF Conference on Computer Vision and Pattern Recognition}, pages 16793--16803, 2022.

\bibitem{zhou2022conditional}
K.~Zhou, J.~Yang, C.~C. Loy, and Z.~Liu.
\newblock Conditional prompt learning for vision-language models.
\newblock In {\em Proceedings of the IEEE/CVF Conference on Computer Vision and Pattern Recognition}, pages 16816--16825, 2022.

\bibitem{zou2023generalized}
X.~Zou, Z.-Y. Dou, J.~Yang, Z.~Gan, L.~Li, C.~Li, X.~Dai, H.~Behl, J.~Wang, L.~Yuan, et~al.
\newblock Generalized decoding for pixel, image, and language.
\newblock In {\em Proceedings of the IEEE/CVF Conference on Computer Vision and Pattern Recognition}, pages 15116--15127, 2023.

\end{thebibliography}
\newpage
\appendix{
\section{Additional Experimental Result}

\paragraph{Training efficiency.} 
To illustrate the training efficiency of \modelname{}, we benchmark its GPU training hours against RegionCLIP and GLIP, as showcased in Table~\ref{tab:training}. 
Even though we utilize the ViT-Large as our backbone, our model achieves faster training. This efficiency can be attributed to our frozen approach and processing of images at a reduced resolution of 224x224 for the CLIP Large.
All benchmarks were executed in a consistent hardware environment, leveraging eight NVIDIA V100 GPUs. 
In stark contrast, GLIP necessitates an extensive 92K GPU hours, a whopping 436 times more than our approach, mainly due to its exhaustive fine-tuning of both Vision and Text models.
Interestingly, even when RegionCLIP adopts a smaller backbone akin to ours, it still requires 4.6K GPU hours.

\begin{table}[h]
    \centering
    \caption{Comparisons with the training efficiency.}
    \small 
    \setlength{\tabcolsep}{4pt} 
    \begin{tabular}{l|c|c|c}
    \hline
    Method & Training data & \begin{tabular}[c]{@{}c@{}}Train time\\ (GPU hours)\end{tabular} & \begin{tabular}[c]{@{}c@{}}Learnable\\ Param (M)\end{tabular} \\
    \hline
    RegionCLIP & CC3M & 4.6K & - \\
    GLIP-T  & O365, GoldG, Cap4M & 92.1K & - \\
    GLIP-L  & FourODs,GoldG,Cap24M & 120K & 289 \\
    GDINO-L  & O365,OI,GoldG,Cap4M,COCO,RefC & no released & 341 \\
    \modelnamepro  & O365, OI, V3D & \textbf{0.2K} & 35 \\
    \hline
    \end{tabular}
    \label{tab:training}
    \vspace{-8pt}
\end{table}

\begin{table}[h]
	\centering
 \vspace{-4pt}
	\begin{center}
        \caption{Effect of increasing the detection training data.} 
        \begin{tabular}{l | c|c|c|c}
       Data & AP$_{r}$ & AP$_{c}$ & AP$_{f}$ & AP  \\
        \shline
        Objects365  & 11.9 & 13.6 & 20.2 & 15.9 \\
        Objects365 + OpenImages   & 16.4 & 18.2 & 23.7 & 20.1 \\
        Objects365 + OpenImages + V3DET  & \textbf{24.9} & \textbf{21.6} & \textbf{25.5} & \textbf{23.7} \\
        \hline\end{tabular}
		\label{tab:data}
		\end{center}
        \vspace{-16pt}
\end{table}

\paragraph{Benefit of increasing the detection training data}
Table~\ref{tab:data} showcases the performance improvements observed when augmenting the training data size. Through our proposed framework, integrating additional detection data from diverse sources consistently enhances the capabilities rooted in pretrained knowledge. 
Compared to training with only Objects365, including OpenImages effectively improves the overall AP from 15.9 to 20.1. The inclusion of V3Det further propels the performance, achieving an impressive overall AP of 23.7. This improvement is particularly significant for rare categories, with an increase from 16.4 to 24.9 (a gain of +8.5 AP), attributable to its extensive vocabulary.

\section{More visualizations on LVIS}
Figure~\ref{fig:more_vis} provides more examples on LVIS~\cite{gupta2019lvis}.

\begin{figure}[t]
    \centering
    \includegraphics[width=0.9\textwidth]{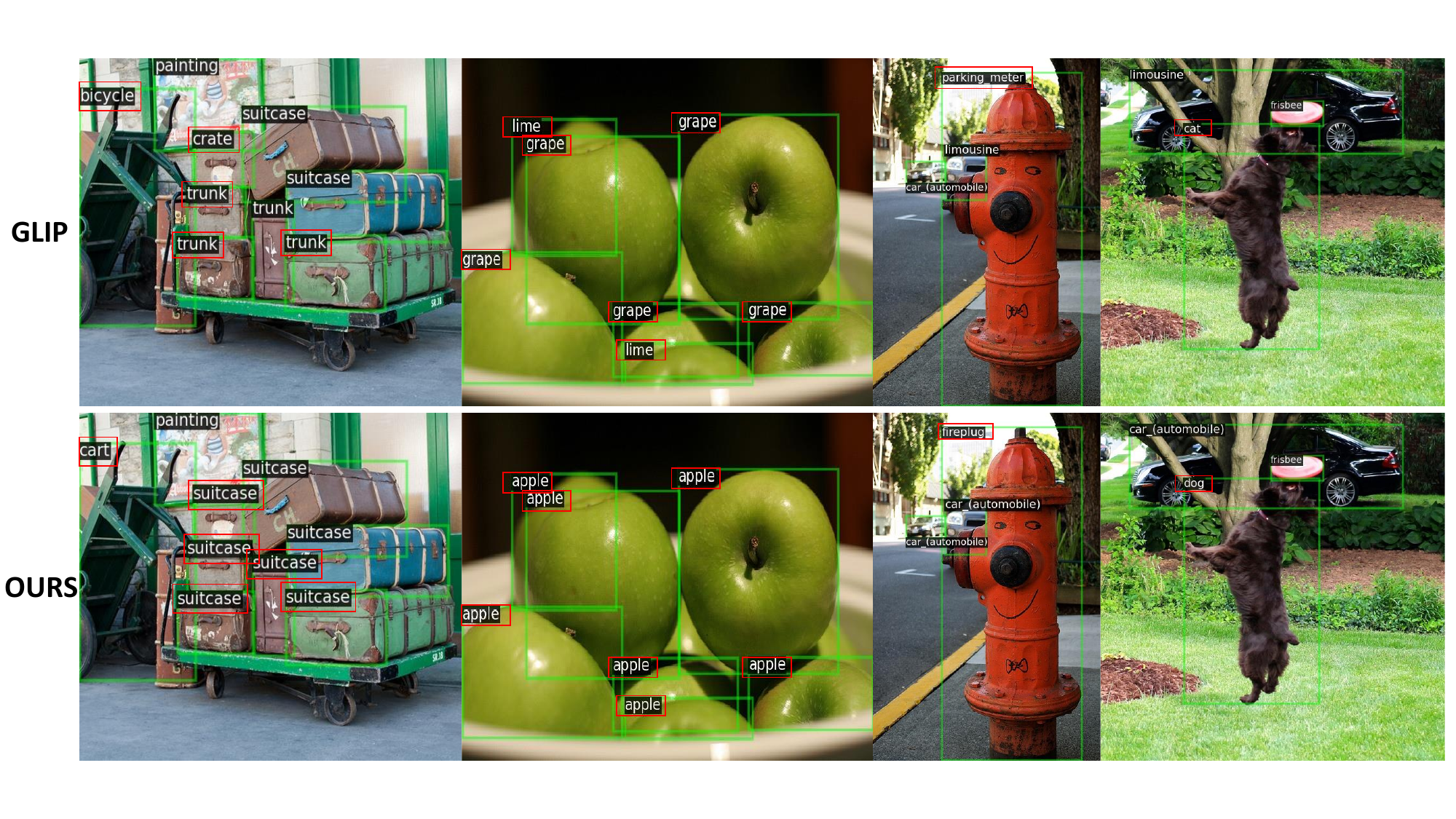}
     \includegraphics[width=0.9\textwidth]{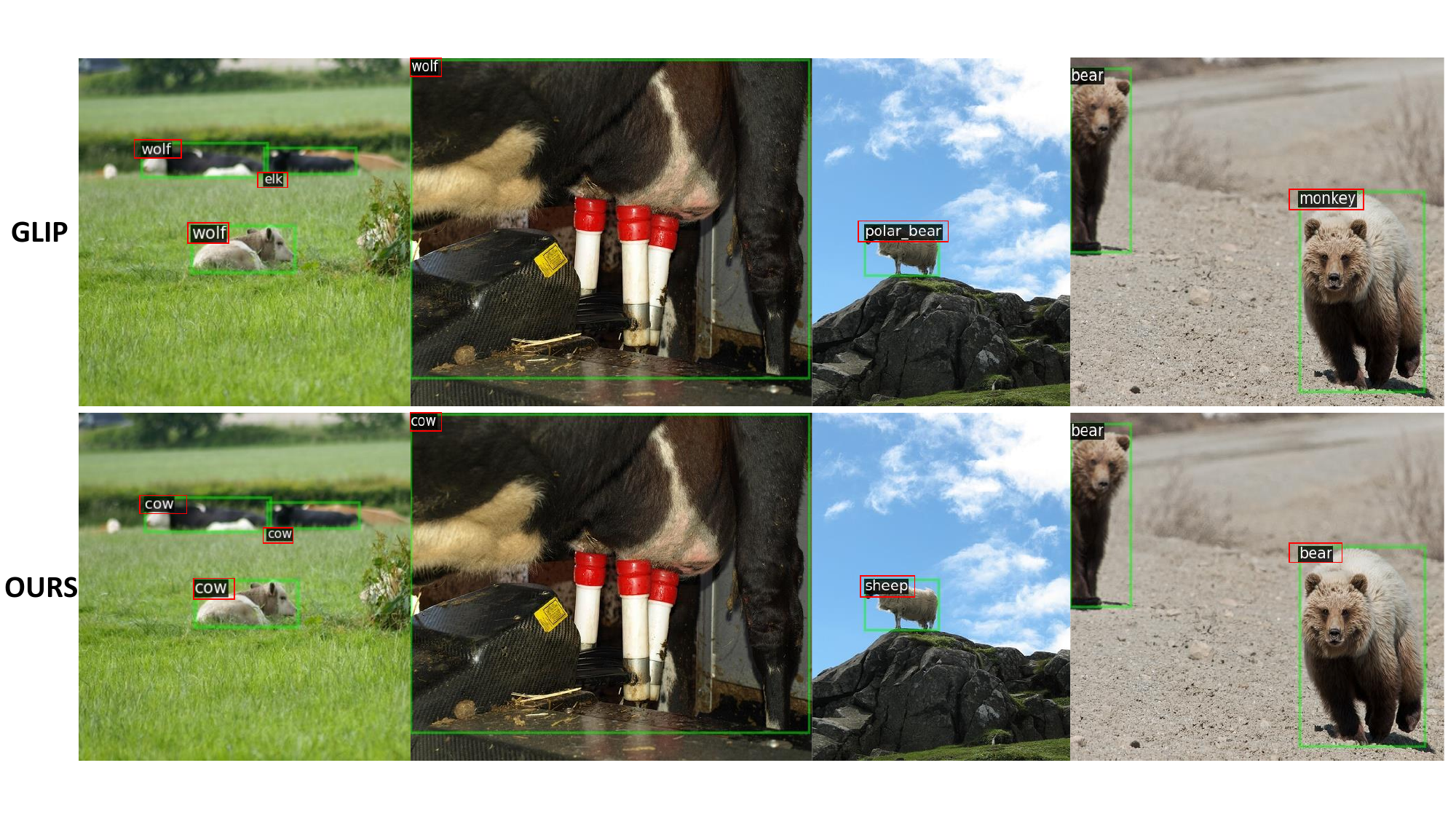}
     \includegraphics[width=0.9\textwidth]{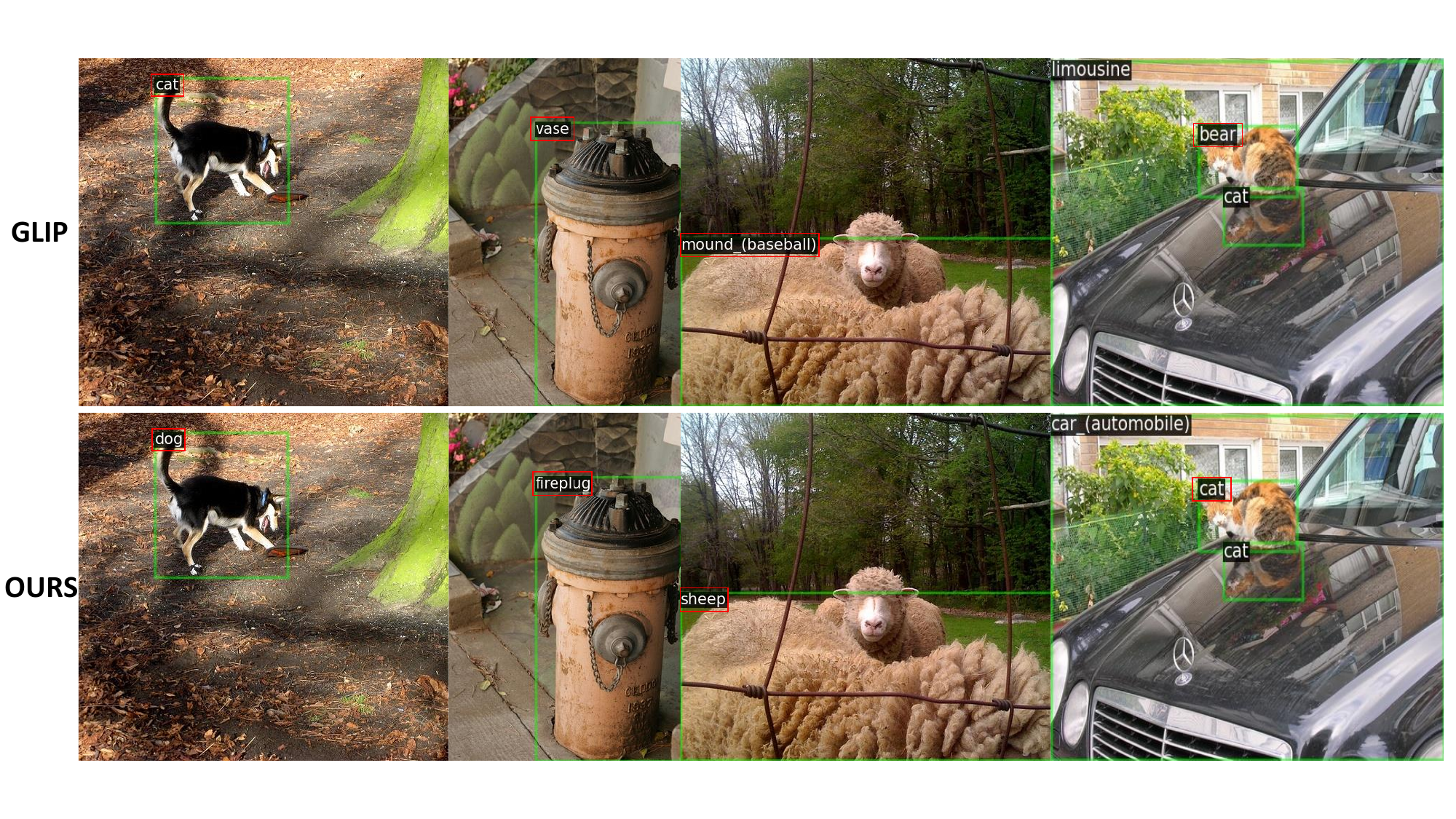}

\caption{
More visualizations in comparison with GLIP. Best viewed when zoomed-in.
}
\label{fig:more_vis}
\end{figure} 
}

\clearpage

\end{document}